\documentclass[letterpaper, 10 pt, conference]{ieeeconf}  

\IEEEoverridecommandlockouts                              

\usepackage{amsmath}
\usepackage{cases}
\usepackage{amssymb}  
\usepackage{amsthm}
\usepackage{algorithm}
\usepackage[noend]{algpseudocode}
\usepackage{subfigure}
\usepackage{graphicx}
\usepackage[utf8]{inputenc}
\usepackage{float}
\usepackage[open,openlevel=1]{bookmark}

\algnewcommand\algorithmicforeach{\textbf{for each}}
\algdef{S}[FOR]{ForEach}[1]{\algorithmicforeach\ #1\ \algorithmicdo}

\makeatletter
\newcommand{\SplitState}[1]{%
  \State
  \parbox[t]{\dimexpr\linewidth-\ALG@thistlm}{%
    #1\par\xdef\Split@prevdepth{\the\prevdepth}%
  }\par
  \prevdepth\Split@prevdepth
}
\makeatother

\DeclareMathOperator{\atantwo}{atan2}

\title{\LARGE \bf
Whole-Body MPC and Dynamic Occlusion Avoidance:\\A Maximum Likelihood Visibility Approach
}

\author {
Ibrahim Ibrahim$^{\dagger}$,
Farbod Farshidian$^{\dagger}$,
Jan Preisig$^{\dagger}$,
Perry Franklin$^{\dagger}$,
Paolo Rocco$^{\ast}$,
Marco Hutter$^{\dagger}$ 
\thanks{This research was supported in part by the Swiss National Science Foundation through the National Centre of Competence in Research Robotics (NCCR Robotics), and in part by TenneT.}%
\thanks{$^\dagger$ Robotic Systems Lab, ETH Z\"{u}rich; $^\ast$ MERLIN, Politecnico di Milano.
}%
}%
        
\begin{document}

\maketitle
\thispagestyle{empty}
\pagestyle{empty}

\begin{abstract}

This paper introduces a novel approach for whole-body motion planning and dynamic occlusion avoidance. The proposed approach reformulates the visibility constraint as a likelihood maximization of visibility probability. In this formulation, we augment the primary cost function of a whole-body model predictive control scheme through a relaxed log barrier function yielding a relaxed log-likelihood maximization formulation of visibility probability. The visibility probability is computed through a probabilistic \emph{shadow field} that quantifies point light source occlusions. We provide the necessary algorithms to obtain such a field for both 2D and 3D cases. We demonstrate 2D implementations of this field in simulation and 3D implementations through real-time hardware experiments. We show that due to the linear complexity of our \emph{shadow field} algorithm to the map size, we can achieve high update rates, which facilitates onboard execution on mobile platforms with limited computational power. Lastly, we evaluate the performance of the proposed MPC reformulation in simulation for a quadrupedal mobile manipulator.

\end{abstract}

\section{Introduction}
A mobile manipulator has a dual advantage of mobility offered by a mobile platform and agility provided by the manipulator. By extending the workspace of fixed-based robotic arms, mobile manipulation systems can access areas far from the reach of the ground-bolted counterparts. However, this mobility comes at the cost of more complex motion planning and navigation problems. Among these challenges, simultaneous motion planning and visual tracking of a target point under inter-object occlusion remain partially unsolved. 

Active vision methods have been a resort in the face of the latter challenge. Those methods often employ sampling in search of the next-best-view (NBV) or in order to perform view planning to maximize the quantity and quality of useful sensory data \cite{b12}. However, active vision methods are still immature in terms of efficiency and efficacy and are still often employed separately from motion planning itself \cite{b7}, \cite{b11}, further increasing the complexity of the workflow.
On the other hand, implementing visual servoing in motion planning applications has limited success in preventing occlusions. In this setting, occlusions are avoided by constraining the camera pixel coordinates of the detected or tracked object of interest. This, in turn, restricts robot operation and control to the camera's Field of View (FOV) \cite{b3}.

At the core of occlusion avoidance lies the visibility problem. Geometrically, visibility is the existence of an unobstructed line of sight between the viewer and the object of interest, and it constitutes a necessary condition for any vision-based task such as detection, tracking, and manipulation. Currently, notions of visibility and obstruction are usually defined deterministically by utilizing ray-casting to simulate the line of sight. However, this approach only evaluates the visibility or lack of it in the scene, which is inadequate for gradient-based techniques and does not reflect reality with all of its uncertainties and sensory noise. The latter often deteriorate the confidence of the target feature tracker and cause drift whenever the line of sight is lost.

To this end, we propose a reformulation of the visibility constraint as a soft constraint in an optimal control setting such that the likelihood of visibility is maximized. This in turn drives the robot to re-establish line of sight whenever it's lost, assuring that the target tracker's confidence is maintained. As such, we are addressing the visibility problem within motion planning itself rather than doing it separately. To complement the reformulated visibility constraint, we also propose a novel method to evaluate notions of visibility and obstruction by building a probabilistic \emph{shadow field} using a Dynamic Programming (DP) approach. This field is constructed using a probabilistic occupancy map computed from RGB-D or LIDAR data and centered around a point light source representing the target of interest that has to be tracked, detected, or manipulated. The key contributions of this paper are:
\begin{itemize}

\item A redefinition of the visibility constraint for optimal control that maximizes the likelihood of visibility in a manner suitable for whole-body motion planning methods.

\item A novel DP-based approach to describe the notions of visibility and obstruction, called \emph{shadow field}, which is inherently probabilistic and efficient to calculate.

\item Hardware mapping tests validating our \emph{shadow field}'s capability in capturing visibility and obstructions.

\item Physics-based simulations verifying our holistic occlusion avoidance implementation within motion planning of a mobile legged manipulator.

\end{itemize}

\section{Related Work}

Visibility constraints, also known as two-dimensional constraints, are often used in Model Predictive Control (MPC) schemes to keep the image-plane coordinates of interest within the camera's FOV \cite{b1,b2,b3}. Such constraints can also represent forbidden regions in the image. However, employing such constraints introduces several challenges. For example, control, in this case, is typically interrupted if the target goes out of FOV \cite{b3}. Mezouar and Chaumette \cite{b17} use a softened visibility constraint which is reformulated as a repulsive potential field, but such a constraint highly increases the chance of having local minima in the overall potential field \cite{b16}.

Occlusion avoidance has been studied across different areas of robotics. In \cite{b5}, Zhang et al. perform dynamic self-occlusion avoidance based on the depth image sequence of moving objects. Objects are first reconstructed then their motion is estimated by matching two Gaussian curvature feature matrices. An optimal planner then decides the camera motion to avoid self-occlusion. In \cite{b13} a prediction method for self-occlusion and mutual occlusion in the case of a multi-arm robotic platform has been proposed based on known geometric models of the objects. Recovery of a UAV target tracker from detection discontinuities caused by occlusion is addressed in \cite{b14} without tackling occlusion in itself. Finally, Nageli et al. \cite{b4} implement occlusion minimization within an MPC scheme. This method, however, is limited to evaluating a single point against an ellipsoidal approximation of obstacles using a fast visibility check. 

Most NBV planning methods are still immature in terms of efficiency and are nowhere near real-time. Wu et al. \cite{b7} perform NBV planning by evaluating the visibility as well as the likelihood of feature matching, achieving around 1.5s per step. On the other hand, \cite{b11} plans the NBV by utilizing online aspect graphs to account for occlusions and feature visibility, requiring more than a minute of planning per movement. NBV methods often utilize ray-casting \cite{b7, b11, b12}. Ray-casting is widely used in robotic planning applications, for example, in autonomous scene exploration \cite{b6} where a dual OcTree structure is used to encode regions which are occupied, free, and unknown, that are then explored via an NBV planning approach. Typically used methods to simulate ray-casting are voxel traversal algorithms such as \cite{b21}.

There exist learning-based methods to evaluate uncertainties arising from image-based detection \cite{b9, b10}. However, such methods are a-posteriori and are incapable of providing meaningful predictive power for optimal control methods. 

The notions of visibility and obstruction are of great interest in multi-agent hide-and-seek or pursuit-evasion applications. In \cite{b19}, a vision cone is constructed for each agent using a LIDAR-like array of 30 rays, and any agent not within the line of sight is masked. Tandon and Karlapalem \cite{b20} represent the notion of visibility in simulation as an associated visibility region for each agent, which itself is also constructed by tracing uniformly spaced rays emitted from the agent. Moreover, the visibility problem is a basic problem in computational geometry \cite{b18, b22, b23} and has applications in computer graphics \cite{b24}.

\section{Problem Formulation}
\label{sec:problem_formulation}

Our goal is to achieve a holistic motion planning formulation for whole-body control of the robot while maximizing the probability of target visibility. Due to the multi-objective nature of our problem, we opt for a finite horizon optimal control formulation, which we solve in an MPC fashion. The optimal control problem is formulated in the continuous-time domain as 
\begin{subequations}
\begin{align}
    & \underset{\boldsymbol{u}(\cdot)}{\text{minimize}} & & \phi(\boldsymbol{x}(t_f)) + \int_{t_s}^{t_f} l(\boldsymbol{x}(t), \boldsymbol{u}(t))dt \label{eq:cost} \\
    &\text{subject to:} & & \boldsymbol{x}(t_s) = \boldsymbol{x}_s \\
    & & & \dot{\boldsymbol{x}} = \boldsymbol{f}(\boldsymbol{x}, \boldsymbol{u}, t) \label{eq:dynamics}\\
    & & &\boldsymbol{g}(\boldsymbol{x}, \boldsymbol{u}, t) = \mathbf{0}  
    \label{eq:equalities} \\
    & & &\boldsymbol{h}(\boldsymbol{x}, \boldsymbol{u}, t) \geq \mathbf{0}, \label{eq:inequalities}
\end{align}
\end{subequations}
where $\boldsymbol{x}$, $\boldsymbol{u}$ are the state and control input respectively. MPC finds the minimum-cost~\eqref{eq:cost} trajectory under the system dynamics $\boldsymbol{f}$, the equality constraints $\boldsymbol{g}$, and the inequality constraints $\boldsymbol{h}$, where $\phi$ is the terminal cost and $l$ is the intermediate cost. In practice, $\boldsymbol{u}(t_s)$ is then applied to control the robot in the receding horizon fashion to allow re-planning with a new measured state, $\boldsymbol{x}_s$. We solve this problem using Sequential Linear-Quadratic (SLQ) solver \cite{farshidian2017efficient} provided by the OCS2 toolbox \cite{OCS2}. We will consider that the visibility and orientation costs, discussed in subsections \ref{subsec:visibility_constraint} and \ref{subsec:orientation_constraint}, act on the robot end-effector (\emph{ee}), but they can generally act on any other frame. We assume that the camera that drives object detection and tracking is mounted on the \emph{ee}.

\subsection{The Reformulated Visibility Constraint}
\label{subsec:visibility_constraint}

Typically, visibility constraints in control problems are enforced by restricting the pixel coordinates of the tracked/ detected object of interest within upper and lower bounds inside the current FOV as shown in \eqref{eq:visibility_inequality}

\begin{equation}
    \left[\begin{array}{c}
        u_{min} \\
        v_{min}
    \end{array}\right] \leq 
    \left[\begin{array}{c}
        u_{act} \\
        v_{act}
    \end{array}\right] \leq
    \left[\begin{array}{c}
        u_{max} \\
        v_{max}
    \end{array}\right].
    \label{eq:visibility_inequality}
\end{equation}

We reformulate those constraints as a maximization of the likelihood that the \emph{ee} has a line of sight to the target within the time-horizon of optimization. The likelihood function, however, is not time-separable, making it unsuitable for trajectory optimization methods. As such, rather than maximizing the likelihood of visibility, we maximize the log-likelihood 
\begin{equation}
    \int_{t_s}^{t_f} \log \mathcal{F}(\boldsymbol{x}_{ee}(t))dt,
    \label{eq:likelihood}
\end{equation} 
where $\mathcal{F}$ in \eqref{eq:likelihood} is the \emph{shadow field} containing information on the visibility probability of the target point, further detailed in section \ref{sec:shadow_field}, and $\boldsymbol{x}_{ee}$ is the 3D position of the \emph{ee} frame of interest within that field. Since we formulated our problem as a minimization, we may augment \eqref{eq:likelihood} within \eqref{eq:cost} as
\begin{equation}
    l_{aug}(\boldsymbol{x}, \boldsymbol{u}) = l(\boldsymbol{x}, \boldsymbol{u}) - \log \mathcal{F}(\boldsymbol{x}_{ee}),
    \label{eq:augmented_cost}
\end{equation} 
where the first term in \eqref{eq:augmented_cost} represents motion planning cost, further detailed in subsection \ref{subsec:motion_planning}, and the second term represents the visibility cost. The gradient of such a cost can be obtained since the gradient of $\boldsymbol{x}_{ee}$ is the Jacobian of the \emph{ee}, and the probabilistic $\emph{shadow field}$ is continuous and smooth, allowing for efficient on-demand trilinear gradient computation at every step.
Since the values returned by $\mathcal{F}(\boldsymbol{x}_{ee})$ can be zero, \eqref{eq:likelihood} is susceptible to extreme values due to the $\log$ function, making it unsuitable for shooting methods. Therefore, we use the relaxed log barrier penalty introduced in \cite{grandia2019feedback}.

\subsection{The Orientation Constraint}
\label{subsec:orientation_constraint}

The reformulated visibility constraint proposed in the previous subsection acts only on the \emph{ee} position, leaving the \emph{ee} orientation unconstrained. As a soft constraint, we add a cost that locks the \emph{ee} onto the target, ensuring that the target remains within the camera's FOV, further enhancing the tracker's confidence about the target. Such a constraint is a function of the emerging solution from the planner along every step of the horizon. $\vec{\mathcal{V}}$ in \eqref{directional_vector} represents the normalized directional vector from the emerging \emph{ee} position $\vec{\mathcal{P}}_{ee}$ to the light position (the target) $\vec{\mathcal{P}}_{L}$. Using $\vec{\mathcal{V}}$, we may extract the yaw and pitch angles in \eqref{yaw} and \eqref{pitch} respectively that orient the \emph{ee} towards the target. The roll angle $\psi$ remains free to choose.
\begin{align}
    \vec{\mathcal{V}} 
    &= \vec{\mathcal{P}}_{L} - \vec{\mathcal{P}}_{ee}
    \label{directional_vector}
    \\
    \phi 
    &= \atantwo \left( \mathcal{V}_{1}, \mathcal{V}_{0} \right)
    \label{yaw}
    \\
    \theta 
    &= -\atantwo \Big(\mathcal{V}_{2},
        \sqrt{{\mathcal{V}_{0}^2} + {\mathcal{V}_{1}^2}}
        \Big).
    \label{pitch}
\end{align}
where $\mathcal{V}_{0}$, $\mathcal{V}_{1}$, and $\mathcal{V}_{2}$ are the components of $\vec{\mathcal{V}}$.

We get the quaternion from those angles, which allows us to compute the normalized quaternion \emph{error} between the desired orientation and the current \emph{ee} orientation. This transformation allows us to safeguard against sudden tracker updates and faults by scaling the constraint's penalty by the factor
\begin{equation}
    \gamma = \frac{
    \max{ \Big(\alpha, \log \big(\frac{\epsilon * error^{\mathsf{c}}}{1 - error^{\mathsf{c}}}\big)\Big)} } {\beta},
    \label{eq:logit}
\end{equation}
where $0 < \alpha < 1$, $\beta \neq 0$, and $\epsilon > 0$ are tuning parameters and $error^{\mathsf{c}}$ is the complement of the normalized quaternion \emph{error}. In this manner, the \emph{ee} gradually and progressively locks onto the target as the quaternion error decreases. We add the cost corresponding to this \emph{error} minimization to $l_{aug}$, constituting the second term of the resulting total cost function as shown in \eqref{eq:total_cost}
\begin{equation}
    l_{total}(\boldsymbol{x}, \boldsymbol{u}) = l_{aug}(\boldsymbol{x}, \boldsymbol{u}) + l_{o}(\boldsymbol{x_{ee}}).
    \label{eq:total_cost}
\end{equation}

\section{Shadow Field}
\label{sec:shadow_field}
\subsection{The Visibility Problem}

\begin{figure}[t]
    \centering
    \includegraphics[width=8.5cm,keepaspectratio]{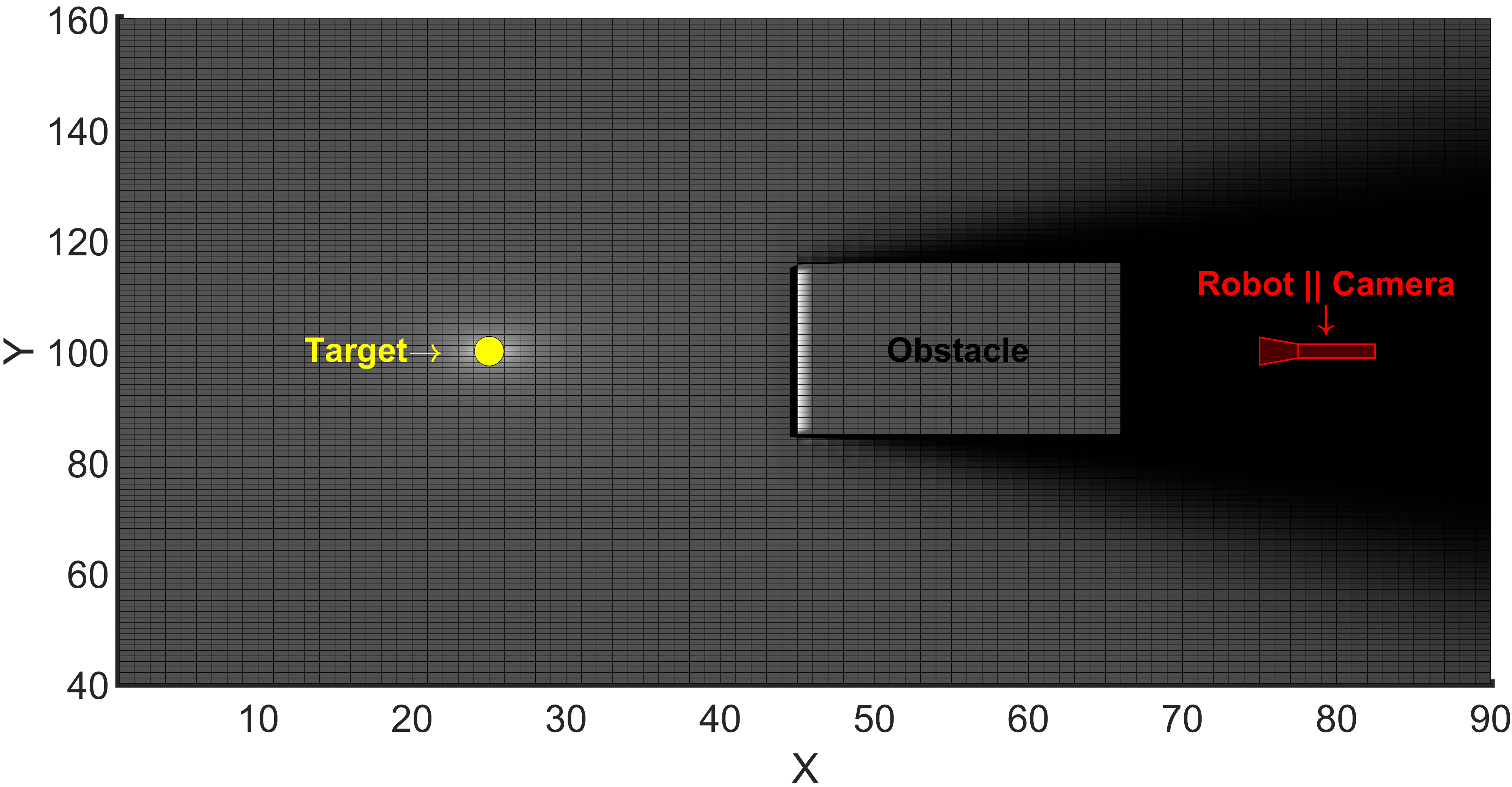}
    \caption{The visibility problem at hand showing an obstacle occluding line-of-sight from a point light source and casting shadows on the robot.}
    \label{fig:visibility_problem}
\end{figure}

The visibility problem at hand is illustrated in Fig.~\ref{fig:visibility_problem}. The tracked target of interest is considered to be a point light source, and a discrete grid represents the field with a fixed resolution. We seek here to calculate \emph{shadow field} which contains the probability of viewing the light source for every pixel of the grid. At the limit of infinite resolution, this map captures a smooth and differentiable field of visibility likelihood between a point and the target. We employ a DP method to compute the map efficiently. In DP, one obtains the solution to a problem by solving simpler sub-problems recursively. Similarly, in our approach, we break down the grid into layers, and we propagate visibility probability solutions starting from the light position, moving outward in every direction one layer at a time. The probability of having a line of sight to the current layer of pixels is used to calculate the likelihood of having a line of sight to the upcoming layer. This procedure is a probabilistic equivalent of ray-casting from the light source to every pixel.

We first introduce the 2D case and then extend the reasoning to the 3D one. In a 2D grid, each pixel in layer $L_{p}$ is affected by two adjacent pixels from the previous layer $L_{p-1}$ which attenuate light based on their own probability of being occupied weighted by a factor that depends on their position relative to the light source. The occupancy probabilities are extracted from a probabilistic occupancy grid whereas the weights constitute a constant environment-independent mapping that only depends on grid resolution.

\begin{figure}[t]
    \centering
    \begin{subfigure}{}
        \includegraphics[width=3.75cm]{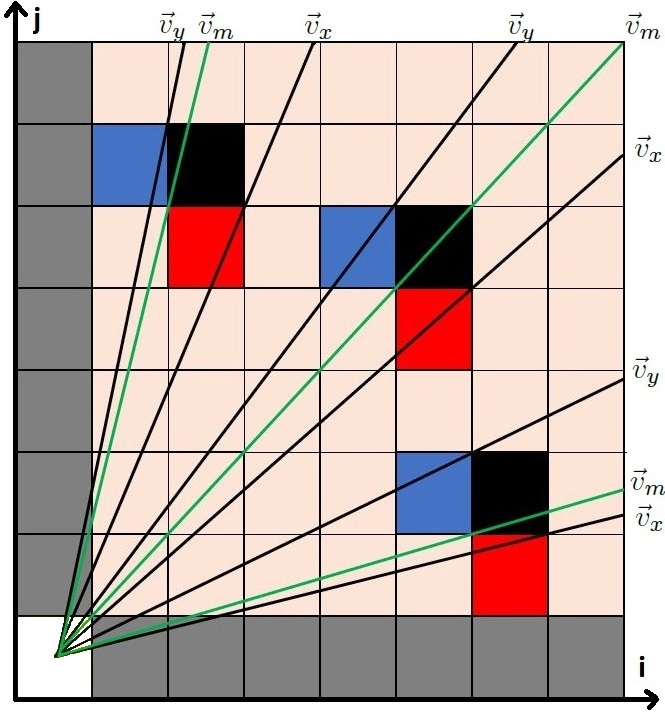}
    \end{subfigure}\hfil
    \begin{subfigure}{}
        \includegraphics[width=4cm]{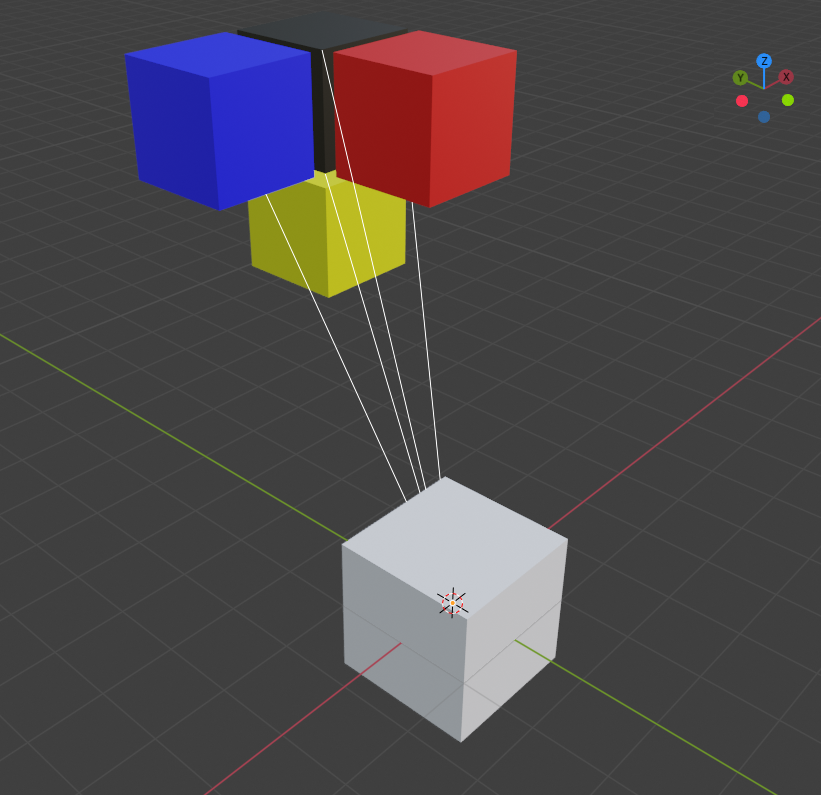}
    \end{subfigure}\hfil
    \caption{Illustrations showing the collapsed 2D visibility problem for 1st quadrant of XY plane (left), and 3D voxels of the light, the queried voxel, and its 3 neighbours in white, black, red, blue and gold (right).}\label{fig:voxel_pixel_llustrations}
\end{figure}

In the first illustration in Fig.~\ref{fig:voxel_pixel_llustrations}, the first quadrant of an XY plane is centered at the light position where three sample queried unoccupied black pixels ($L_{p}$) and their corresponding neighboring red and blue pixels ($L_{p-1}$) are displayed. For each black pixel, two black vectors passing through $(i+1,j)$ and $(i,j+1)$ and a green one passing through $(i,j)$ are highlighted, referred to as $\vec{v}_{x}$, $\vec{v}_{y}$, $\vec{v}_{m}$ respectively, where $(i,j)$ is the lower left corner of the black pixel. The two angles $\widehat{\vec{v}_{x} \vec{v}_{m}}$ and $\widehat{\vec{v}_{y} \vec{v}_{m}}$, and their sum $\widehat{\vec{v}_{x} \vec{v}_{y}}$, are computed using trigonometry. The contributions of every red $(i,j-1)$ and blue $(i-1,j)$ pixel to light attenuation are set equal to their probabilities of being occupied weighted by the complementary ratios $\widehat{\vec{v}_{x} \vec{v}_{m}}$ to $\widehat{\vec{v}_{x} \vec{v}_{y}}$ and $\widehat{\vec{v}_{y} \vec{v}_{m}}$ to $\widehat{\vec{v}_{x} \vec{v}_{y}}$ respectively. Referring to the same illustration, one may notice that as the black pixel moves further along x relative to y or vice versa, one angle progressively dominates the other, meaning that the light will be progressively attenuated by one pixel and regressively by the other. If the black pixel is occupied, however, its visibility probability is set to the complement of its occupancy probability.



\begin{figure}[t]
    \centering
    \includegraphics[width=8cm,keepaspectratio]{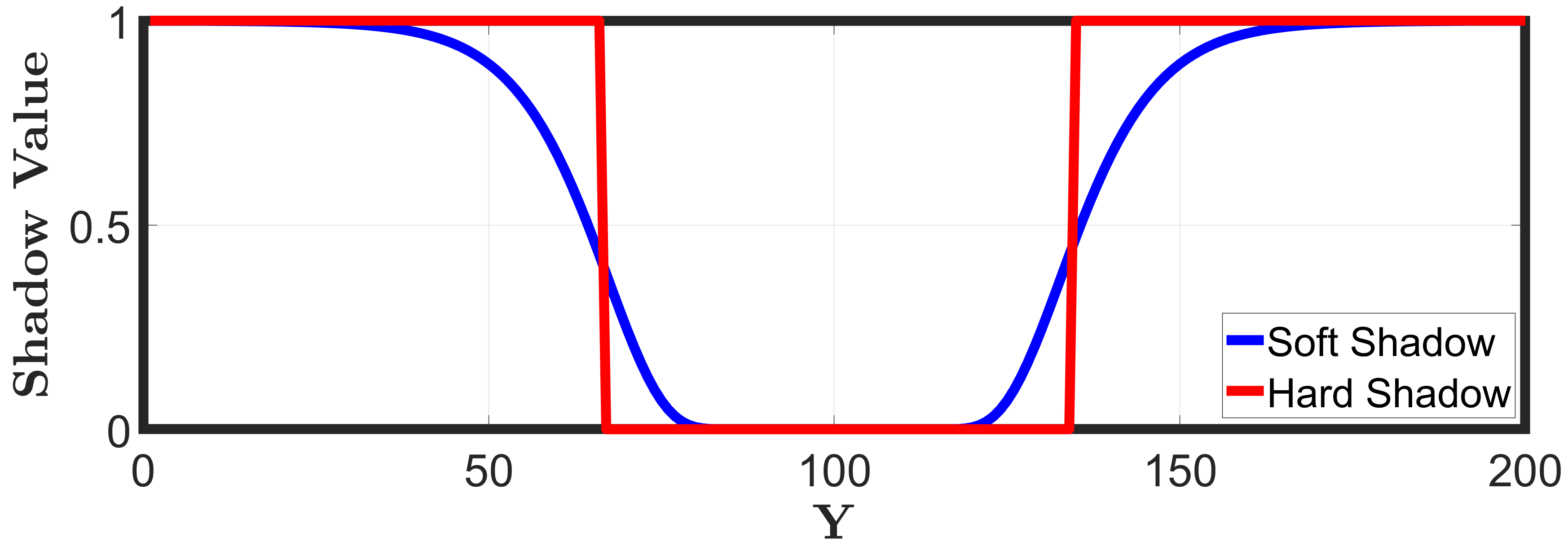}
    \caption{Values of our soft \emph{shadow field} approximation against hard shadow values produced by the obstacle in Fig.~\ref{fig:visibility_problem} along X = 70.}
    \label{fig:SoftVsHard}
\end{figure}

In Fig.~\ref{fig:SoftVsHard}, we compare hard shadow versus our soft shadow values evaluated along the vertical line $X = 70$ for the scene in Fig.~\ref{fig:visibility_problem}. This comparison highlights the extent and accuracy of our soft shadow approximation relative to hard shadows. Owing to its probabilistic nature and in contrast to a hard shadow technique, our \emph{shadow field} provides a smooth representation of visibility with a value ranging from 0, representing cold regions, to 1, representing hot regions. Colder regions are regions lying in the shadow of the light source, i.e., it is less likely that a line of sight exists between those regions and the light source. On the contrary, hotter regions are regions where a line of sight to the light is more likely to exist. The highlighted smoothness plays an essential role in applications where the field's gradient is required, such as our MPC formulation.

\begin{figure*}[t]
    \centering
    \begin{subfigure}{}
        \includegraphics[width=5.725cm]{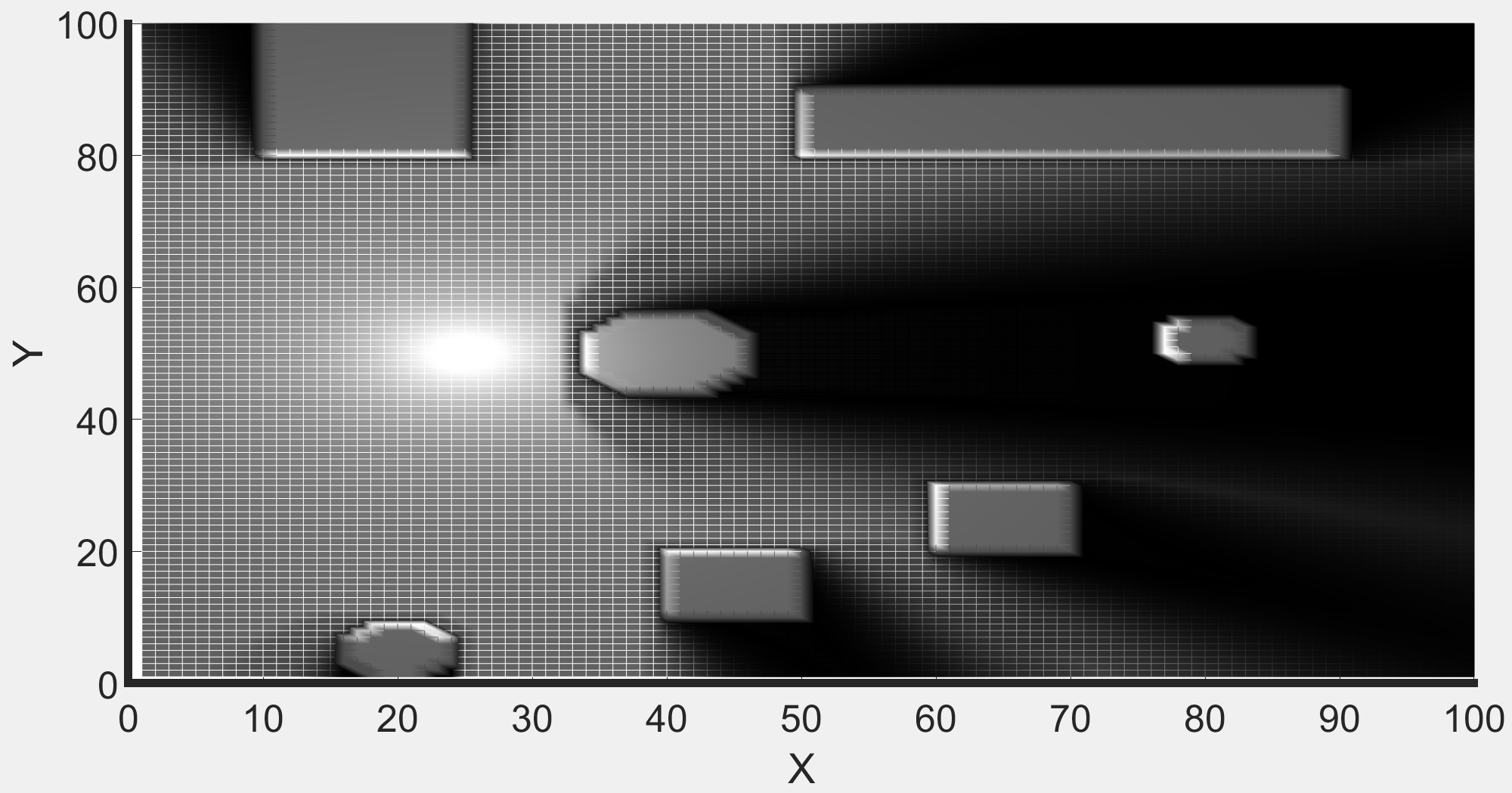}
    \end{subfigure}\hfil
    \begin{subfigure}{}
        \includegraphics[width=5.725cm]{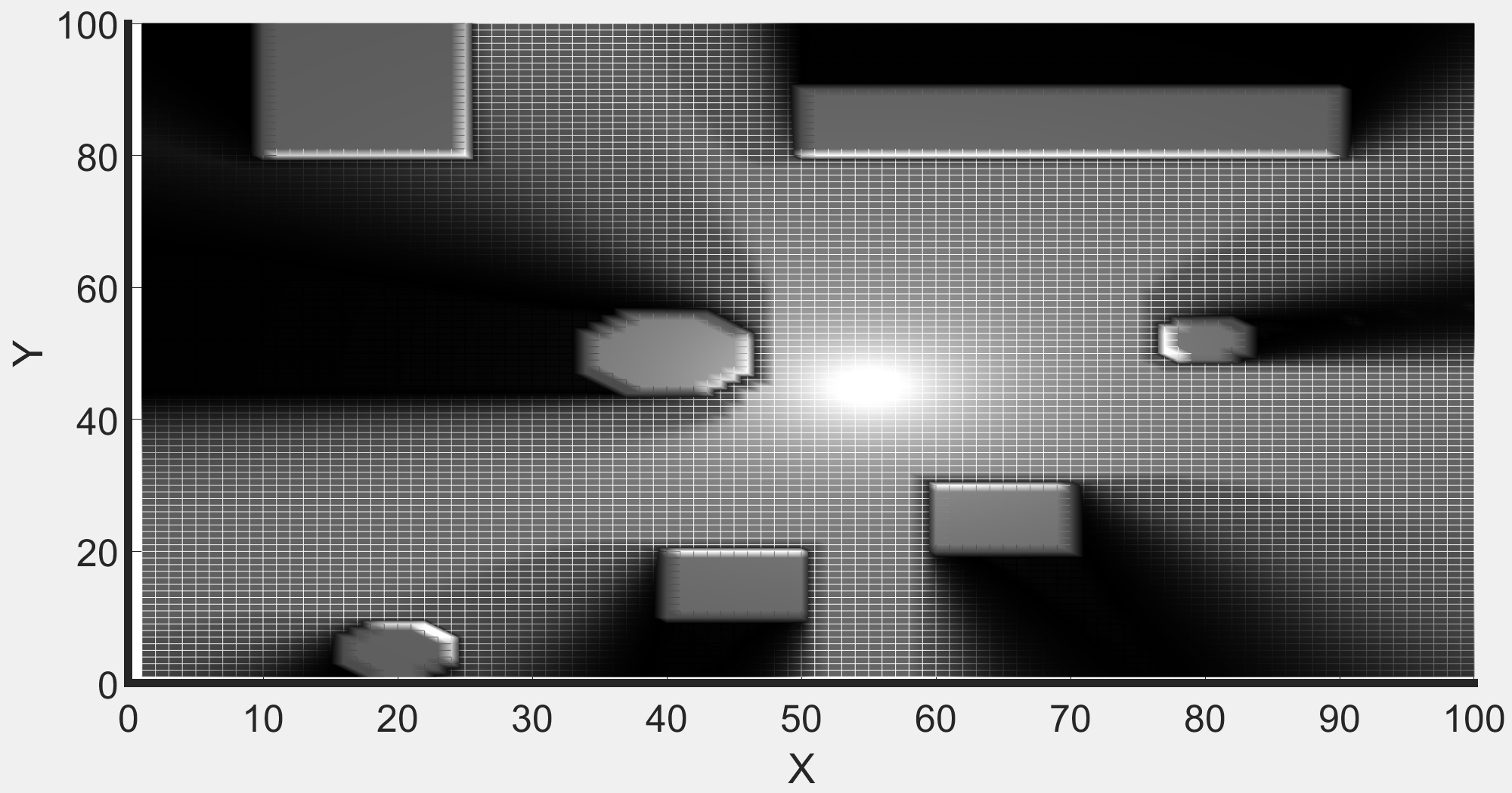}
    \end{subfigure}\hfil
    \begin{subfigure}{}
        \includegraphics[width=5.725cm]{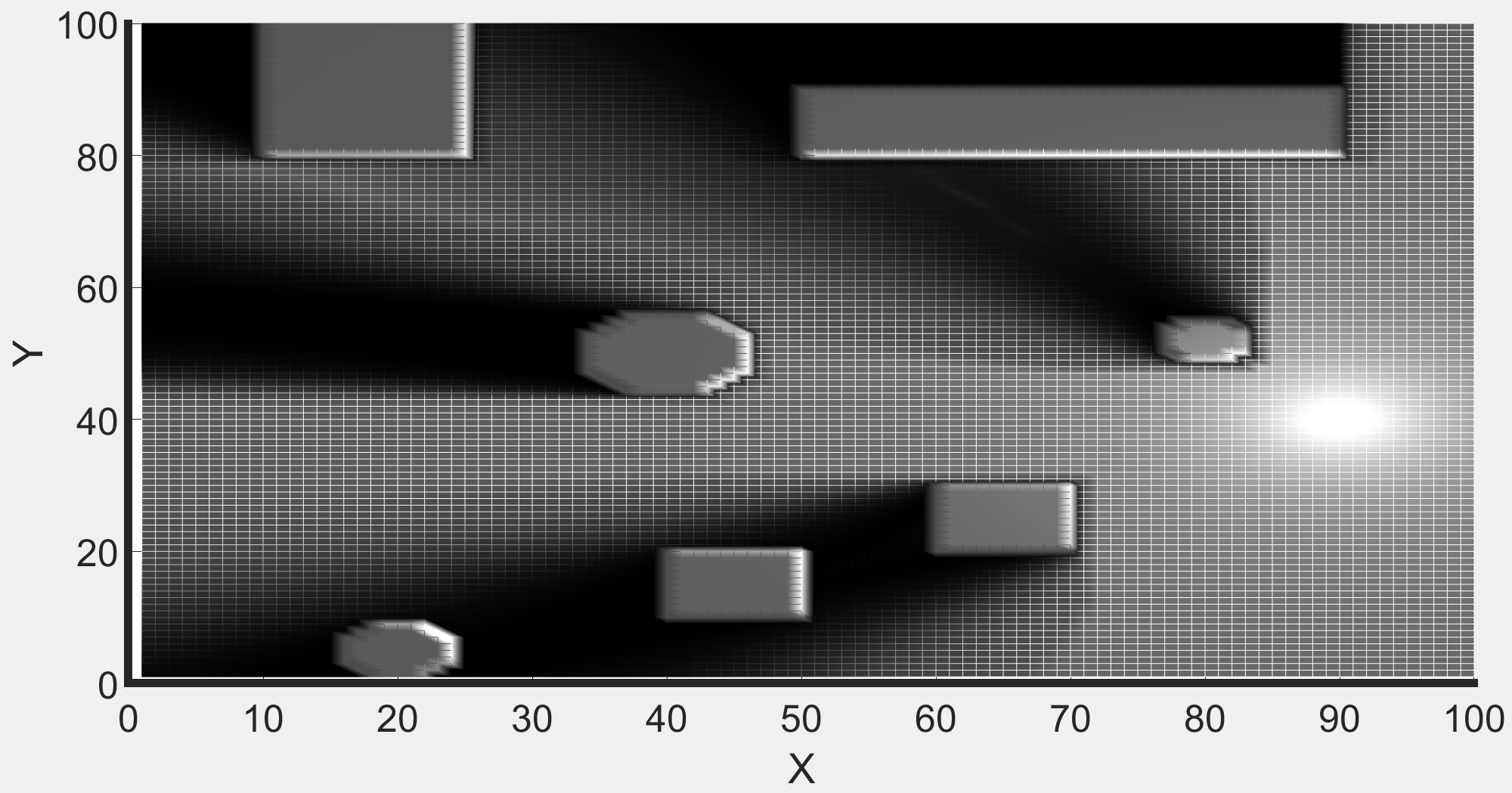}
    \end{subfigure}\hfil
    \begin{subfigure}{}
        \includegraphics[width=5.725cm]{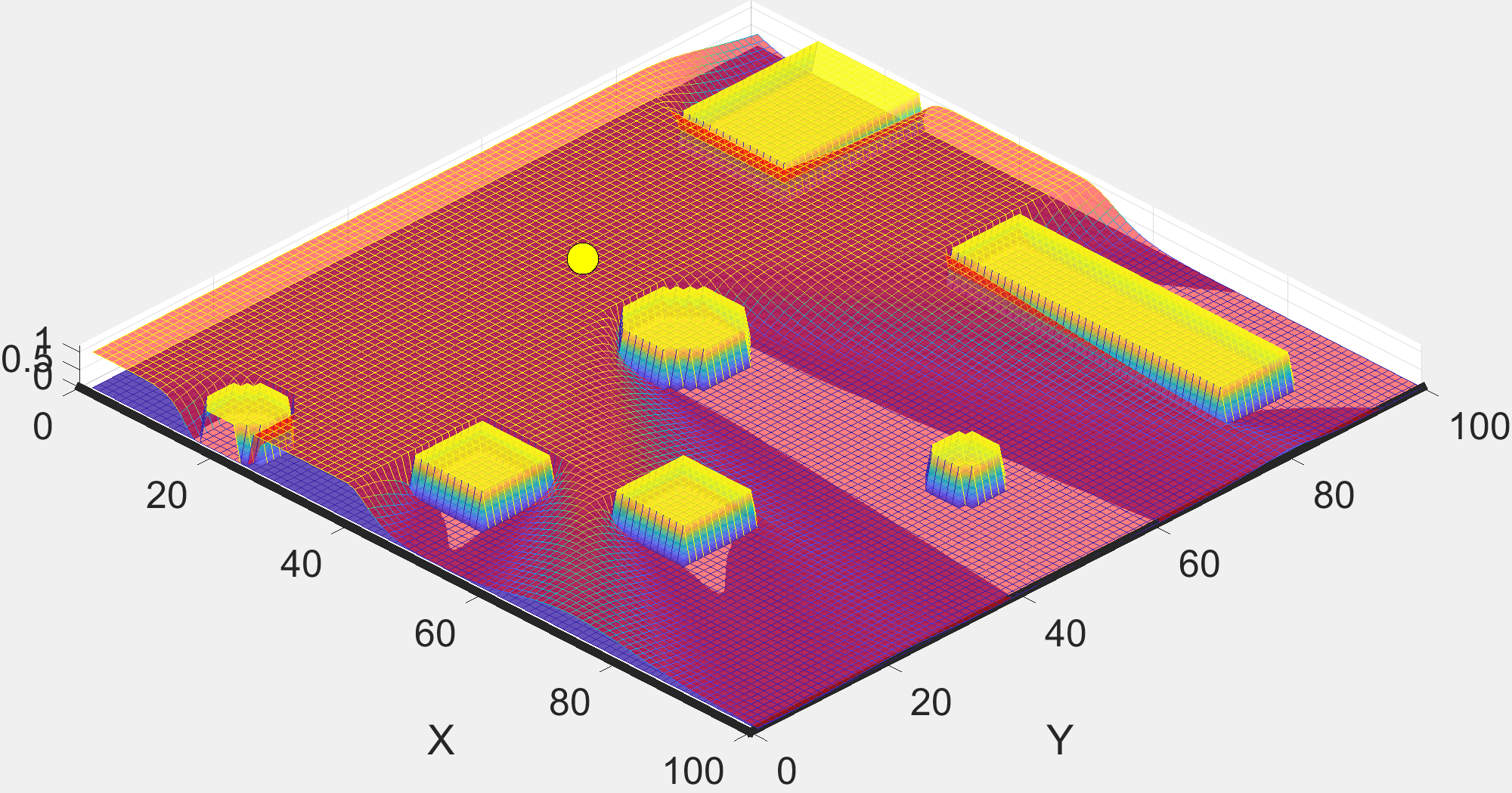}
    \end{subfigure}\hfil
    \begin{subfigure}{}
        \includegraphics[width=5.725cm]{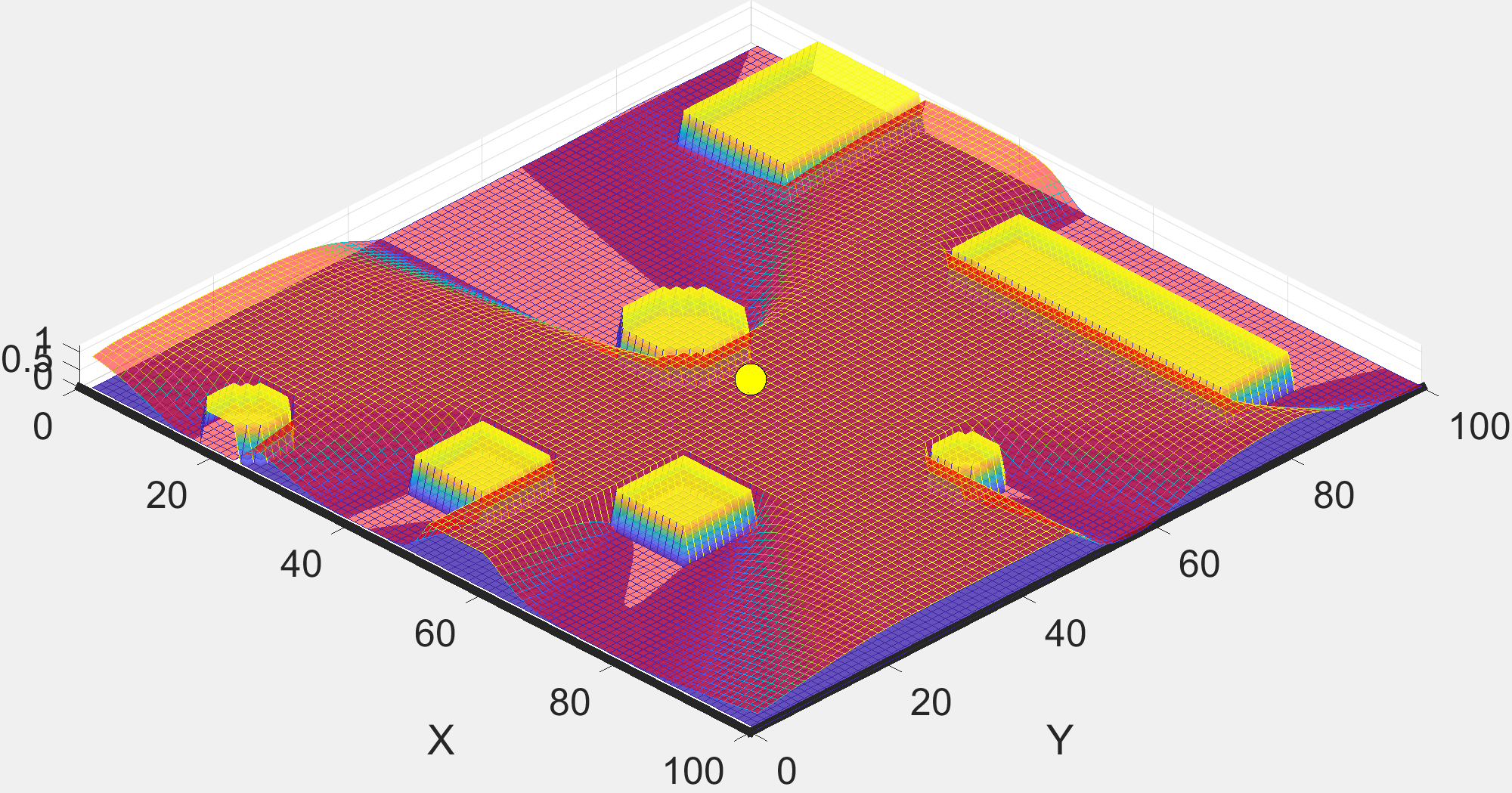}
    \end{subfigure}\hfil
    \begin{subfigure}{}
        \includegraphics[width=5.725cm]{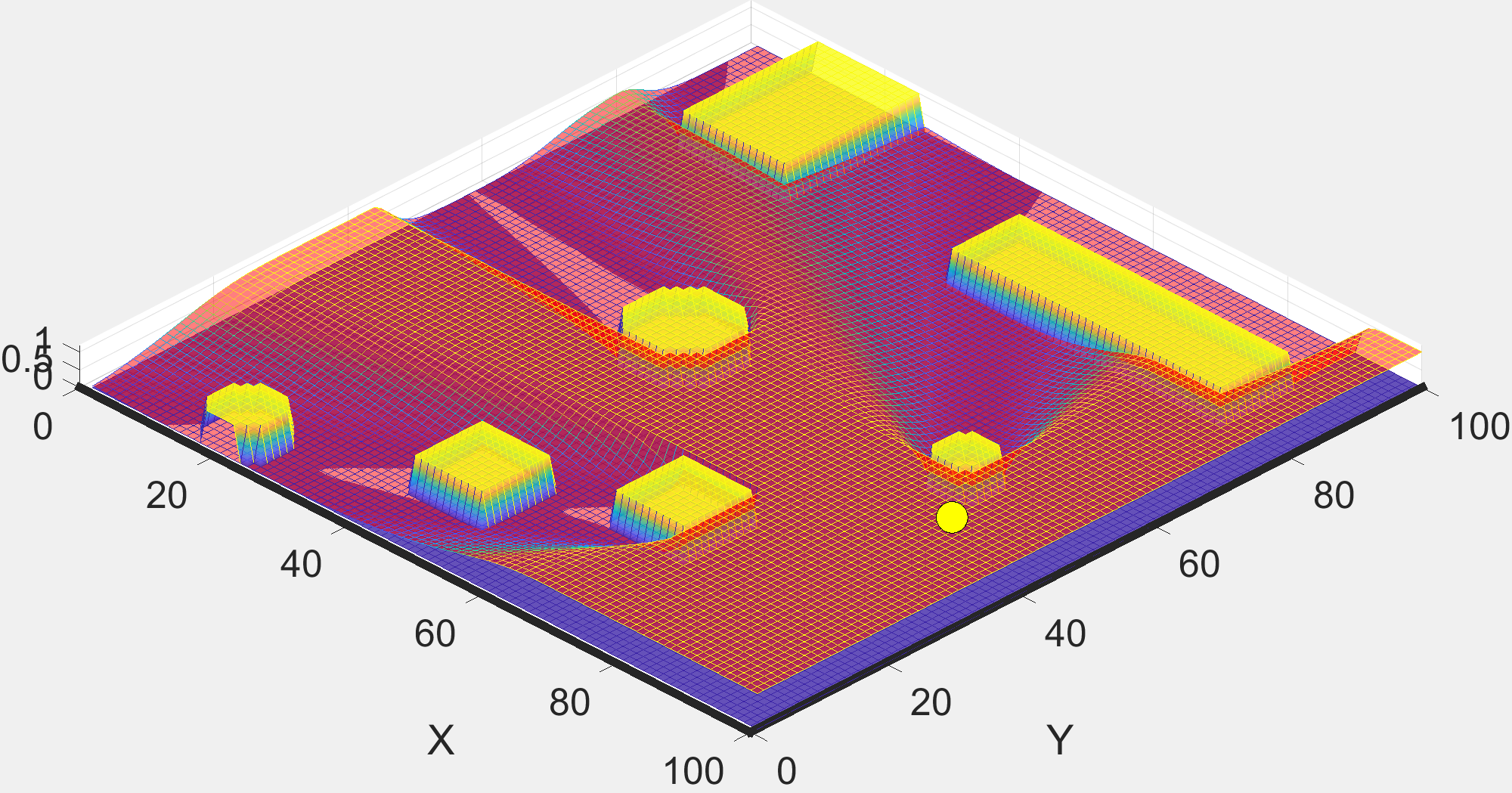}
    \end{subfigure}\hfil
    \caption{Three pairs of one top-view and its corresponding side-view of a 2D \emph{shadow field} that results from obstacles (yellow face color) occluding a point light source. The environment is identical in all illustrations, the only difference being the light position.}\label{fig:2D_Animation}
\end{figure*}

We display in Fig.~\ref{fig:2D_Animation} three sample meshes representing the \emph{shadow field} of a 2D environment grid of size $100\times100$ at two different views as we vary the light position. The generated wavy \emph{shadow field} meshes in the side-views are characterized by the same smoothness as the soft shadow curvature shown in Fig.~\ref{fig:SoftVsHard} while both umbras and penumbras are produced for each occluding obstacle as presented in the top-views.

\subsection{3D Weights and Update Algorithms}
We extend the same reasoning from 2D to 3D since the 3D scenario is an extension of the 2D case, where in addition to the XY plane one needs to consider the XZ and YZ planes. Each queried voxel in $L_{p}$ is affected by exactly three neighbouring ones in $L_{p-1}$. An example of a queried 3D voxel ($L_{p}$) with its three neighbours ($L_{p-1}$) is shown in the second illustration in Fig.~\ref{fig:voxel_pixel_llustrations} as black, red, blue, and gold respectively, whereas the light position is the white one. Also, four white vectors are shown to be extending from the light position to locations $(x_{i}, y_{i}, z_{i})$, $(x_{i+1}, y_{i}, z_{i})$, $(x_{i}, y_{i+1}, z_{i})$, and $(x_{i}, y_{i}, z_{i+1})$, where $(x_{i}, y_{i}, z_{i})$ is the black cube's vertex nearest to the light source. Those vectors are referred to $\vec{v}_{m}$, $\vec{v}_{x}$, $\vec{v}_{y}$, and $\vec{v}_{z}$ in Alg.~\ref{alg:Algorithm 1}. The normals to the planes formed by vectors $\vec{v}_{x}$ and $\vec{v}_{y}$, $\vec{v}_{x}$ and $\vec{v}_{z}$, and $\vec{v}_{y}$ and $\vec{v}_{z}$ are then computed. Using those normals, the angles between vector $\vec{v}_{m}$ and the three aforementioned planes as well as their sum may be obtained. In the final step, the respective weights distributions for every voxel are calculated and stored in the mappings $\mathcal{W}_{r}$, $\mathcal{W}_{b}$, and $\mathcal{W}_{g}$. 

\begin{algorithm}[t]
    \small
    \caption{Caching Shadow Field Weights} \label{alg:Algorithm 1}
    \begin{algorithmic}[1]
        \Procedure{initShadowFieldWeights()}{}
        \State$\mathcal T\gets$  empty volume of size$(x, y, z)$
        \State Initialize $\mathcal{W}_{r}$, $\mathcal{W}_{b}$, and $\mathcal{W}_{g}$ to $\mathcal T$
        \ForAll {$z_{i} \text{ in z } \in \mathcal T$}
            \ForAll {$x_{i} \text{ in x } \in \mathcal T$}
                \ForAll {$y_{i} \text{ in y } \in \mathcal T$}
                    \SplitState {$\vec{v}_m \gets (x_{i}, y_{i}, z_{i})$ \\ $\vec{v}_x \gets (x_{i+1}, y_{i}, z_{i})$ \\ $\vec{v}_y \gets (x_{i}, y_{i+1}, z_{i})$\\ $\vec{v}_z \gets (x_{i}, y_{i}, z_{i+1})$}
                    \SplitState {
                    $\vec{n}_{xy}\gets \vec{v}_{y} \times \vec{v}_{x}$ \\
                    $\vec{n}_{xz}\gets \vec{v}_{x} \times \vec{v}_{z}$ \\
                    $\vec{n}_{yz}\gets \vec{v}_{z} \times \vec{v}_{y}$}
                    \SplitState {
                    $a_{xy} \gets \arcsin{ \left( \frac{\vec{v}_m \cdot \vec{n}_{xy}}{ \lVert \vec{v}_{m} \rVert \cdot \lVert \vec{n}_{xy} \rVert } \right)}$ \\ 
                    $a_{xz} \gets \arcsin{ \left( \frac{\vec{v}_m \cdot \vec{n}_{xz}}{ \lVert \vec{v}_{m} \rVert \cdot \lVert \vec{n}_{xz} \rVert } \right)}$ \\ 
                    $a_{yz} \gets \arcsin{ \left( \frac{\vec{v}_m \cdot \vec{n}_{yz}}{ \lVert \vec{v}_{m} \rVert \cdot \lVert \vec{n}_{yz} \rVert } \right)}$ \\
                    $a_{sum}\gets \left( a_{xy} + a_{xz} + a_{yz} \right)$ 
                    }
                    \SplitState {$\mathcal{W}_{r}(x_{i}, y_{i}, z_{i})\gets \frac{a_{yz}}{a_{sum}}$ \\
                    $\mathcal{W}_{b}(x_{i}, y_{i}, z_{i})\gets \frac{a_{xz}}{a_{sum}}$ \\
                    $\mathcal{W}_{g}(x_{i}, y_{i}, z_{i})\gets \frac{a_{xy}}{a_{sum}}$}
                \EndFor
            \EndFor
        \EndFor
        \State \textbf{return} $\mathcal{W}_{r}$, $\mathcal{W}_{b}$, $\mathcal{W}_{g}$
        \EndProcedure
    \end{algorithmic}
\end{algorithm}

\begin{algorithm}[t]
    \small
    \caption{Updating Shadow Field}\label{alg:Algorithm 2}
    \textbf{Inputs:} $\mathcal L \gets$ volume $(x_{local}, y_{local}, z_{local})$ valued \textit{1.0} \\
    \hspace*{10mm} $\mathcal{O} \gets$  probabilistic occupancy grid of the global map \\
    \hspace*{10mm} $(l_{{g}_{x}}, l_{{g}_{y}}, l_{{g}_{z}}) \gets$  light indices in global map \\
    \hspace*{10mm} $(l_{{l}_{x}}, l_{{l}_{y}}, l_{{l}_{z}}) \gets$  light indices in local map \\
    \hspace*{10mm} $(x_{p}, y_{p}, z_{p}) \gets$ \# of indices till upper boundary of $\mathcal L$ \\
    \hspace*{10mm} $(x_{n}, y_{n}, z_{n}) \gets$ \# of indices till lower boundary of $\mathcal L$ \\
    \hspace*{10mm} $\mathcal{P}_{{\mathcal{O}}_{thresh}} \gets$ confidence level for occupancy \\
    \hspace*{10mm} $\mathcal{W}_{r}, \mathcal{W}_{b}, \mathcal{W}_{g} \gets$ cached weights \\
    \textbf{Output: } $ \mathcal{F} \gets$ 3D probabilistic field
    \begin{algorithmic}[1]
        \Procedure{updateShadowField()}{} 
        \State \textbf{Initialize} \textit{$\mathcal{F}$ to $\mathcal L$ }
        \For {$z_{i} = 0 \text{ to $z_p$ }$, $x_{i} = 0 \text{ to $x_p$ }$, $y_{i} = 0 \text{ to $y_p$}$,}
                \State $(x_{{l}_{i}}, y_{{l}_{i}}, z_{{l}_{i}}) = (l_{{l}_{x}}, l_{{l}_{y}}, l_{{l}_{z}}) + (x_{i}, y_{i}, z_{i}) $
                \State $(x_{{g}_{i}}, y_{{g}_{i}}, z_{{g}_{i}})= (l_{{g}_{x}}, l_{{g}_{y}}, l_{{g}_{z}}) + (x_{i}, y_{i}, z_{i})$
                \State $\mathcal{P}_{\mathcal{O}} \gets \mathcal{O}(x_{{g}_{i}}, y_{{g}_{i}}, z_{{g}_{i}})$
                \If{$\mathcal{P}_{\mathcal{O}} > P_{{\mathcal{O}}_{thresh}}$}
                    \State $\mathcal{F}(x_{{l}_{i}}, y_{{l}_{i}}, z_{{l}_{i}}) \gets (1 - P_{\mathcal{O}})$
                \Else
                    \SplitState {$\mathcal{C}_{r} \gets \mathcal{W}_{r}(x_{i}, y_{i}, z_{i}) * \mathcal{F}(x_{{l}_{i-1}}, y_{{l}_{i}}, z_{{l}_{i}})$ \\ 
                    $\mathcal{C}_{b} \gets \mathcal{W}_{b}(x_{i}, y_{i}, z_{i}) * \mathcal{F}(x_{{l}_{i}}, y_{{l}_{i-1}}, z_{{l}_{i}})$ \\
                    $\mathcal{C}_{g} \gets \mathcal{W}_{g}(x_{i}, y_{i}, z_{i}) * \mathcal{F}(x_{{l}_{i}}, y_{{l}_{i}}, z_{{l}_{i-1}})$ \\ 
                    $\mathcal{F}(x_{{l}_{i}}, y_{{l}_{i}}, z_{{l}_{i}}) \gets \mathcal{C}_{r} + \mathcal{C}_{b} + \mathcal{C}_{g}$}
                \EndIf
        \EndFor
        \State \textbf{return} $F$
        \EndProcedure
    \end{algorithmic}
\end{algorithm}

\emph{Shadow field} constitutes a local subset of the global map and is centered around the light source, with a maximum number of indices separating between the light source index and upper and lower bounds of the \emph{shadow field}, $(x_{p}, y_{p}, z_{p})$ and $(x_{n}, y_{n}, z_{n})$ respectively. Alg.~\ref{alg:Algorithm 2} takes those numbers as input, as well as the probabilistic occupancy grid $\mathcal{O}$, the indices corresponding to the global and local light positions $(l_{{l}_{x}}, l_{{l}_{y}}, l_{{l}_{z}})$ and $(l_{{g}_{x}}, l_{{g}_{y}}, l_{{g}_{z}})$, the cached weights $\mathcal{W}_{r}$, $\mathcal{W}_{b}$, $\mathcal{W}_{g}$, and a tunable confidence level for the probability of being occupied $\mathcal{P}_{{\mathcal{O}}_{thresh}}$. Alg.~\ref{alg:Algorithm 2} outputs the \emph{shadow field}. Alg.~\ref{alg:Algorithm 2} updates all cells in the field one quadrant at a time, starting from the light source position. For every voxel in level $L_{p}$, if its occupancy probability $\mathcal{{P}_{O}}$ is higher than the threshold $\mathcal{P}_{{\mathcal{O}}_{thresh}}$, then assign to it a value of 1 - $\mathcal{{P}_{O}}$, otherwise update it based on contributions of neighbouring red, blue, and golden voxels from $L_{p-1}$: $\mathcal{C}_{r}$, $\mathcal{C}_{b}$, and $\mathcal{C}_{g}$. 
For brevity, we only present the first quadrant update formula, since each of the other quadrants requires only few sign changes in the iteration logic. 

\subsection{Computational and Storage Complexity}

Referring to Alg.~\ref{alg:Algorithm 2}, the upper bound of our run-time computational complexity is 2 additions and 3 multiplications per voxel, which scales linearly with the number of voxels and cubically with the grid resolution. Alg.~\ref{alg:Algorithm 1} scales similarly. This is very useful, especially when running other demanding computational processes. Since our method requires no special hardware, it may run on a separate CPU. This offloads computation from the main processor allowing it to perform demanding motion planning for a complex system with high degrees of freedom which is the case for our experiments, as shall be further discussed in section \ref{sec:results}. The storage complexity for the 3D case, excluding the occupancy grid, is that of the \emph{shadow field} and the mappings $\mathcal{W}_{r}$, $\mathcal{W}_{b}$, and $\mathcal{W}_{g}$, which is exactly $4 ( n_x\times n_y\times n_z)$.

\section{Results}
\label{sec:results}
\begin{figure}[t]
    \centering
    \includegraphics[width=4cm,keepaspectratio]{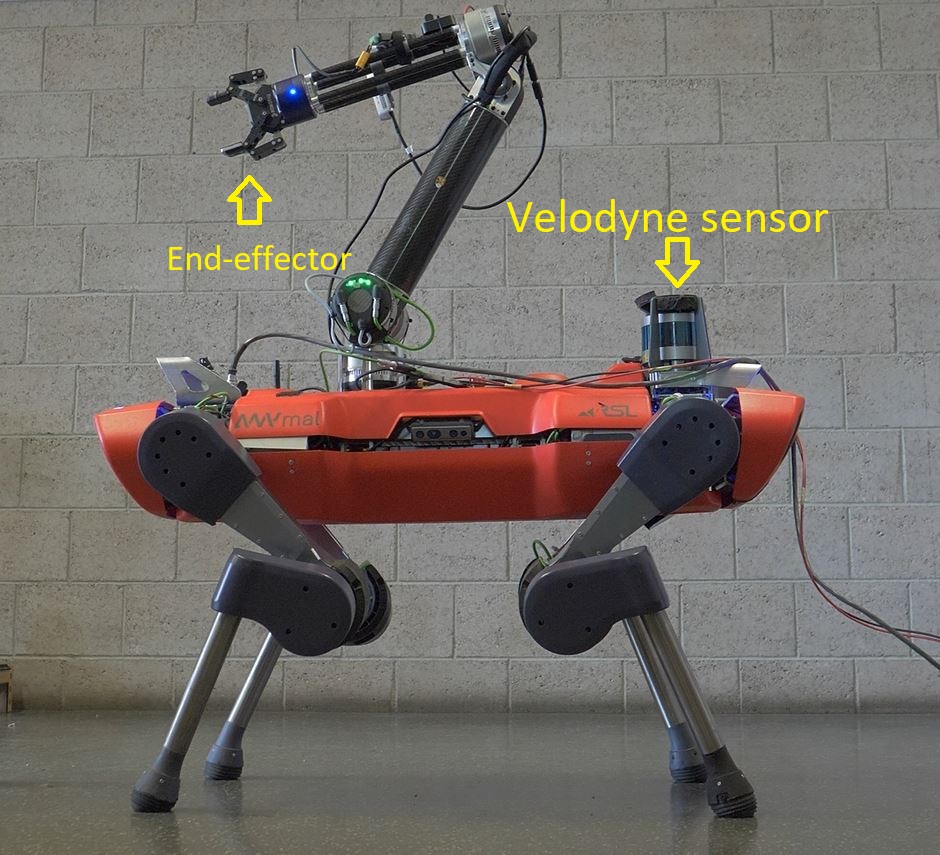}
    \caption{ALMA, a robot composed of a custom-made, torque-controllable 4 DoF arm, DynaArm, mounted on a quadrupedal platform, ANYmal C. ALMA is equipped with four Realsense D435 cameras that provide $360^{\circ}$ vision as well as a VLP16 Puck LITE Velodyne sensor mounted on its rear.}
    \label{fig:Alma}
\end{figure}

\begin{figure}[t]
    \centering
    \begin{subfigure}{}
        \includegraphics[width=4cm]{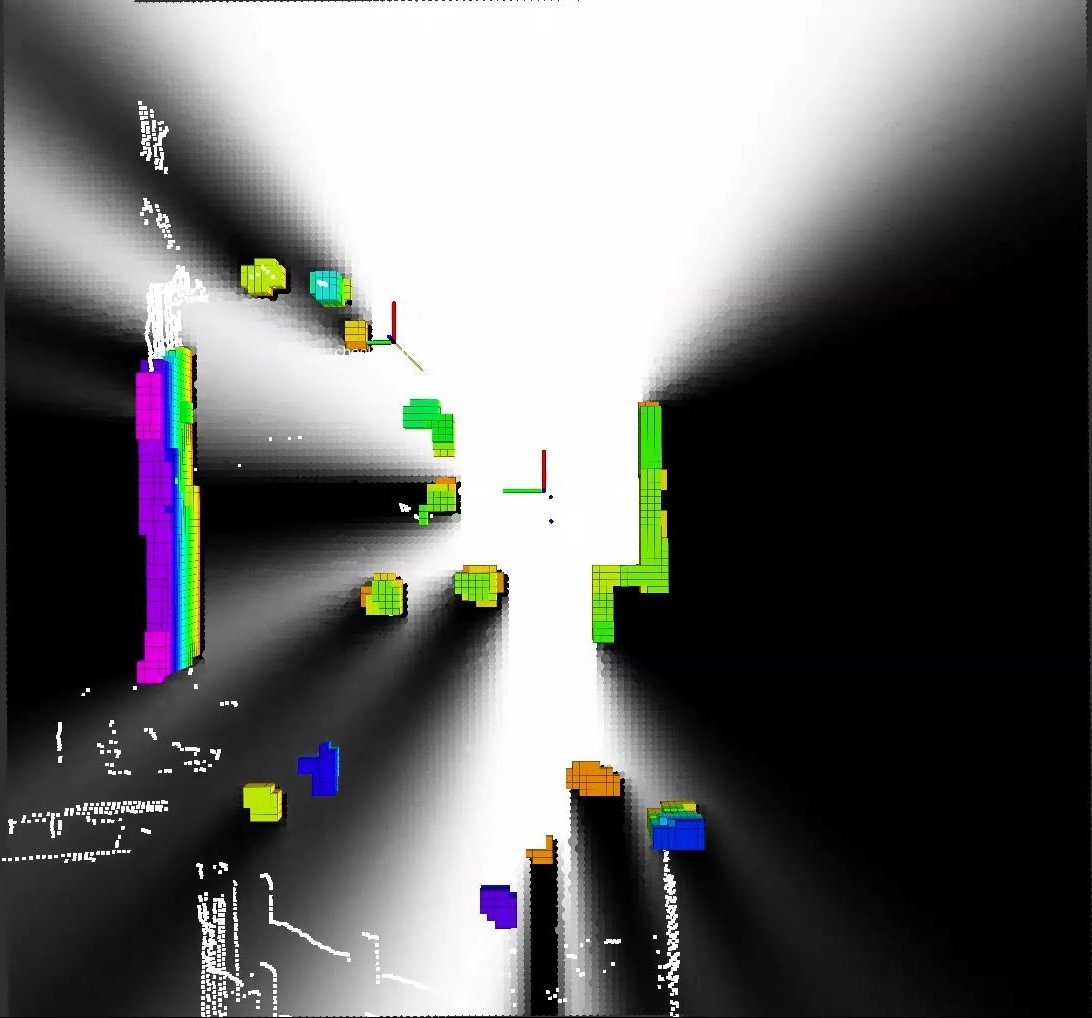}
    \end{subfigure}\hfil
    \begin{subfigure}{}
        \includegraphics[width=3.95cm]{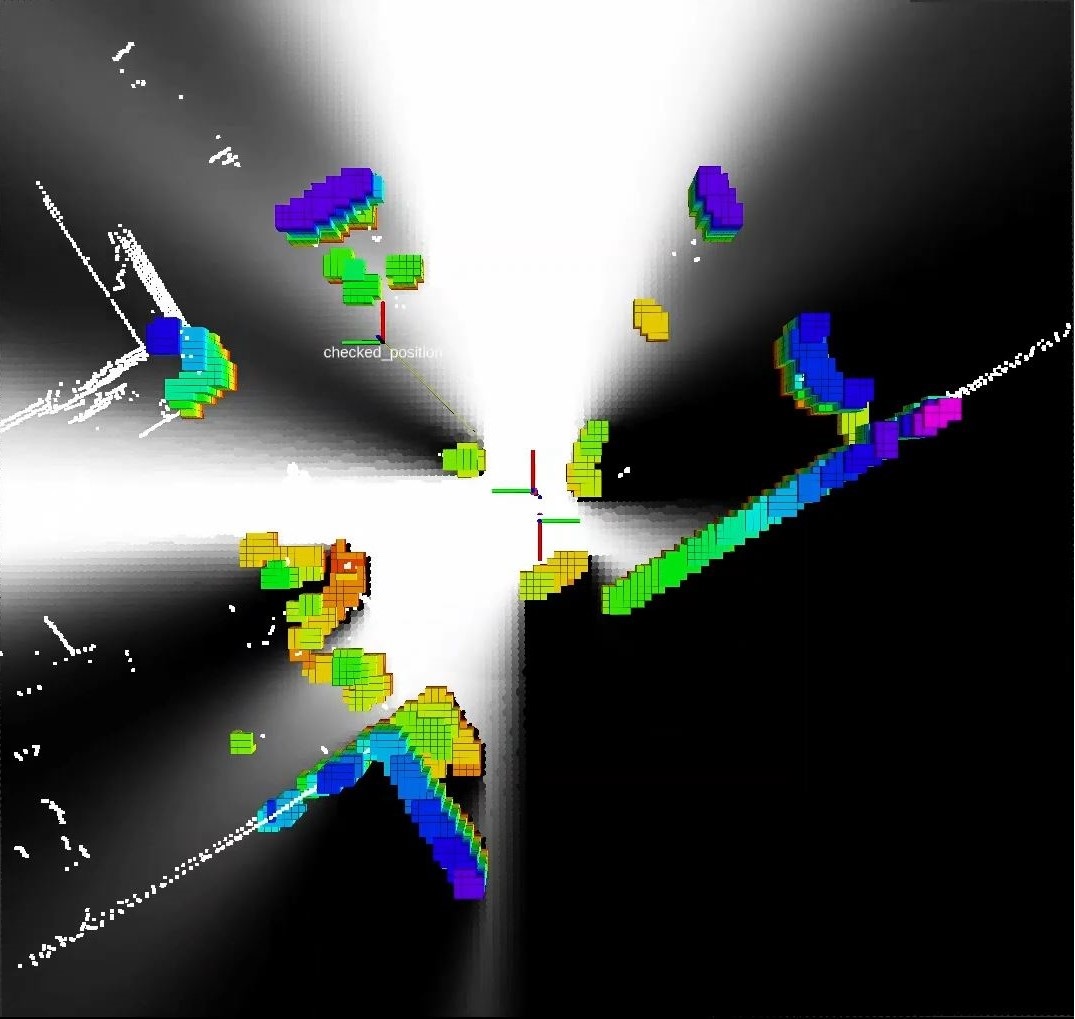}
    \end{subfigure}\hfil
    \caption{Top view of the occupied voxels (colored), the Velodyne pointcloud (small white dots), and the resulting \emph{shadow field} pointcloud slice (gray-scale) from the real-time hardware mapping. The light source and the \emph{ee} are at the center of each image. The umbras produced by the occluding voxels are the pitch black regions while their penumbras are the shades of gray.}\label{fig:shadow_field_hw_test}
\end{figure}

In this section, we validate the proposed \emph{shadow field} on real data from hardware and demonstrate the performance of the complete control pipeline in simulation on a quadrupedal mobile manipulator, ALMA \cite{b25}, Fig.~\ref{fig:Alma}. It has two onboard computers, one running locomotion and planning modules, and the other running mapping modules. The adopted planning framework in our physics-based experiments is based on the whole-body motion planner formulation introduced in \cite{b25}. We then augment our costs as discussed in section III. Our results can be seen in the provided video\footnote{https://www.youtube.com/watch?v=qDOYlOT8G-4}.

\subsection{Real-time Hardware Shadow Field Mapping}

To validate our \emph{shadow field}, two real-time mapping experiments are considered. In each of them, ALMA is placed in a different cluttered environment, and data from the Veldoyne sensor are used to construct a 3D probabilistic occupancy grid. The occupancy grid is then used to compute a 3D \emph{shadow field} spanning $16\times16\times2$ m$^{3}$ at a resolution of 1000 voxels per cubed meter. The onboard mapping computer runs the whole pipeline at rates exceeding 100Hz, far beyond the Velodyne pointcloud update rate ($\sim$15Hz). For visualization purposes, the whole 3D \emph{shadow field} is computed, but only a horizontal slice of it is visualized at the \emph{ee} level in the form of a pointcloud having gray-scale intensities proportional to the values of the \emph{shadow field}. At that slice level, brighter areas represent visible regions while darker areas represent occluded regions. In both experiments, we coincide the light position with the \emph{ee} position at the center of the field, i.e., we are evaluating the \emph{ee}'s sight of the surrounding scene. We also visualize the occupied voxels, which have different colors correlated to their occupation probability. We show the top view of the slice, the occupied voxels, and the Velodyne pointcloud in Fig.~\ref{fig:shadow_field_hw_test}. The resulting shadows retain the same smoothness and continuity characteristics as the soft shadow presented in Fig.~\ref{fig:SoftVsHard}, plus the occupancy grid's noise.

\subsection{Motion Planning and Control Pipeline}
\label{subsec:motion_planning}

\begin{figure}[t]
    \centering
    \begin{subfigure}{}
        \includegraphics[width=4cm]{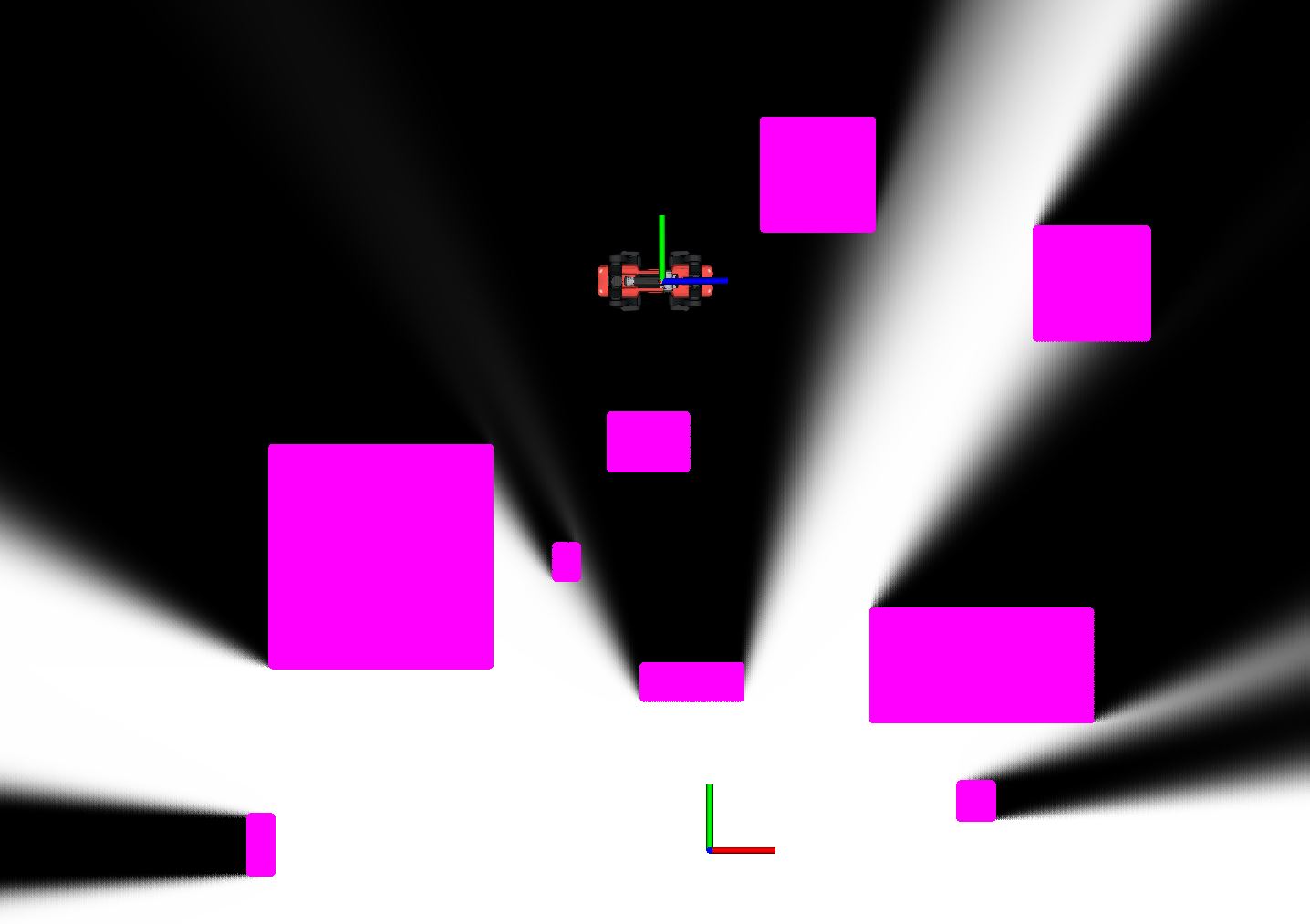}
    \end{subfigure}\hfil
    \begin{subfigure}{}
        \includegraphics[width=4cm]{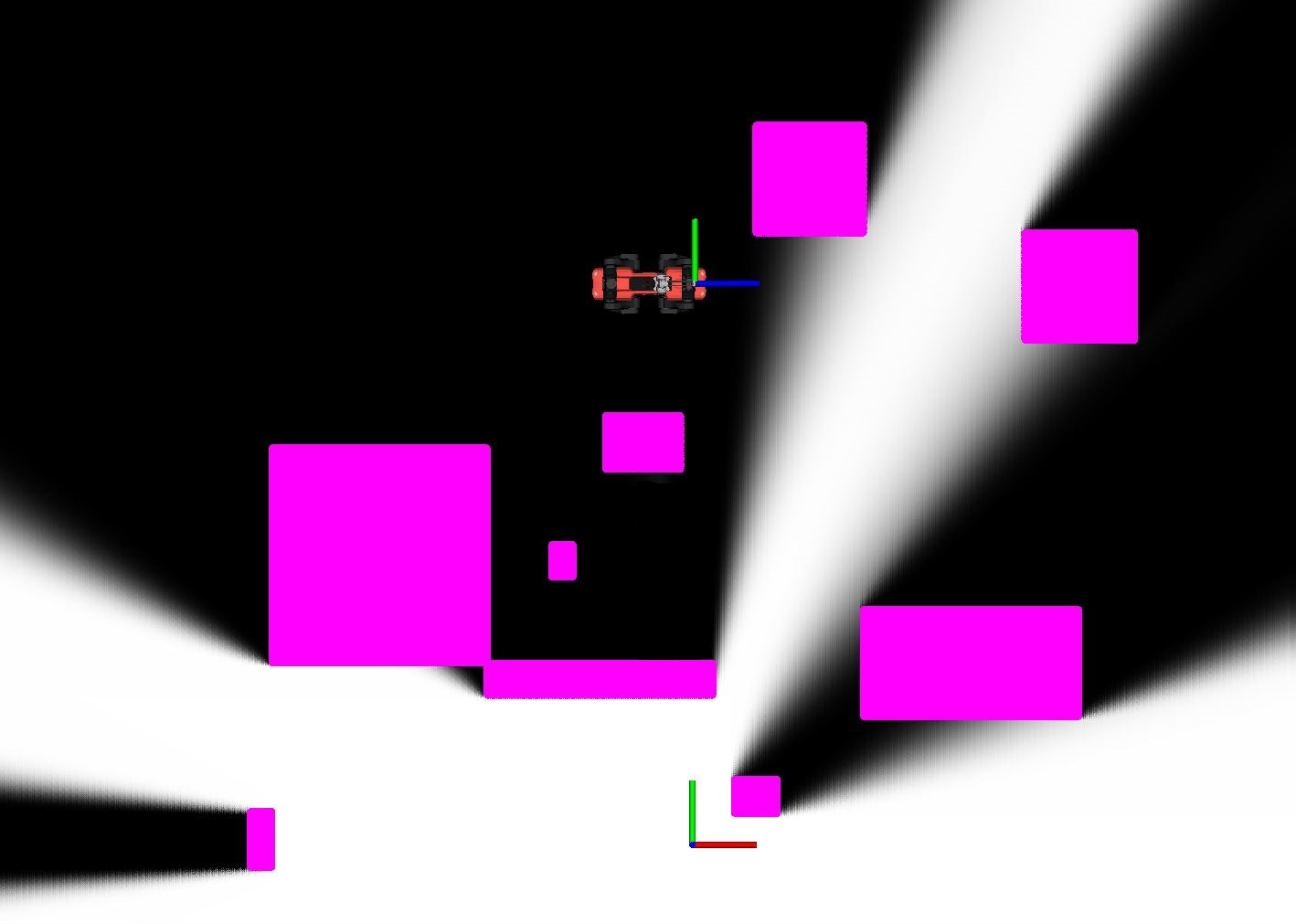}
    \end{subfigure}\hfil
    \caption{Simulation scenes containing obstacles shown in violet, ALMA shown as the top frame, and the light source shown as the bottom frame.}\label{fig:sim_scenes}
\end{figure}

\begin{figure}[t]
    \centering
    \begin{subfigure}{}
        \includegraphics[width=8cm]{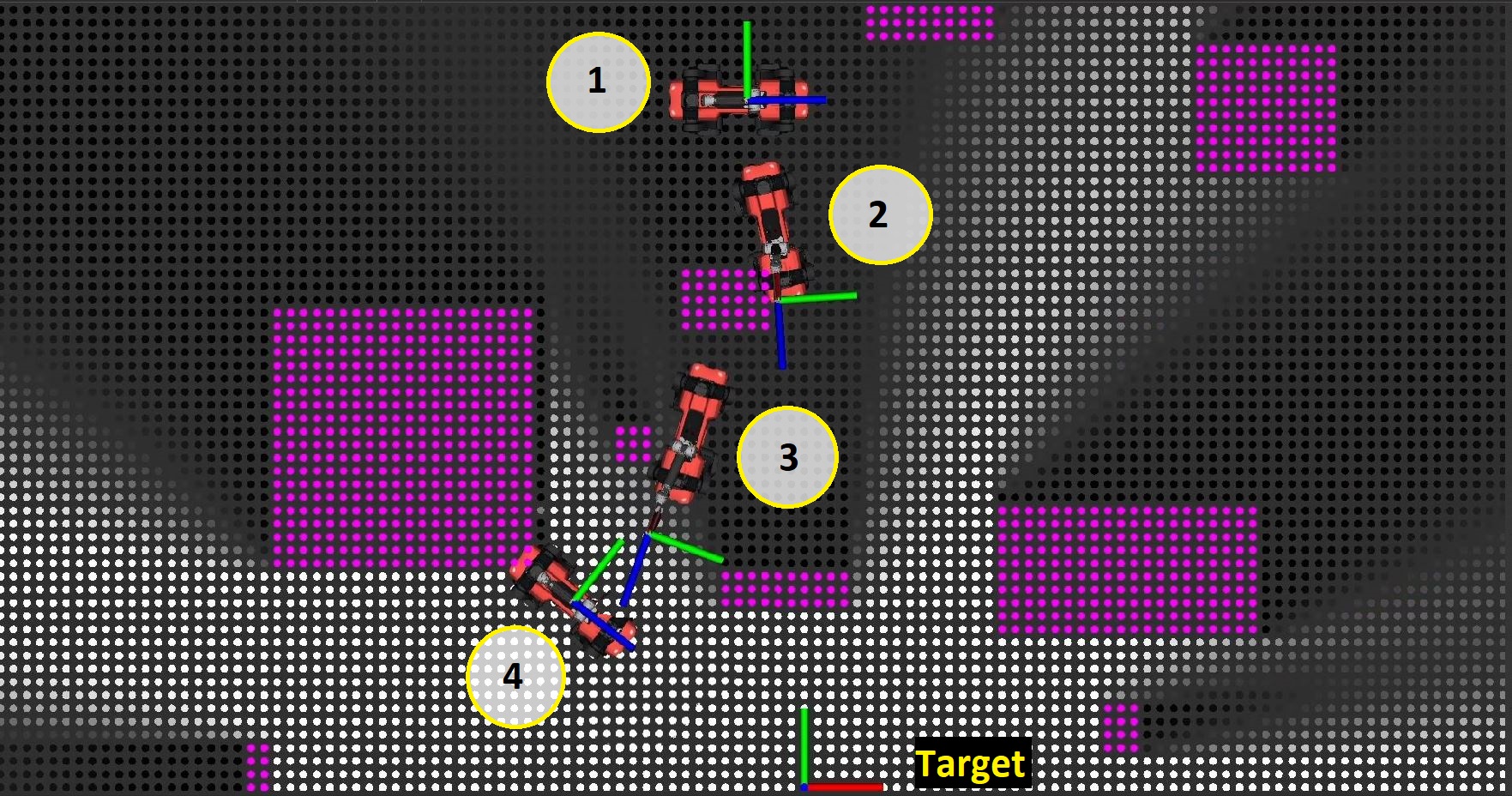}
    \end{subfigure}\hfil
    \begin{subfigure}{}
        \includegraphics[width=8cm]{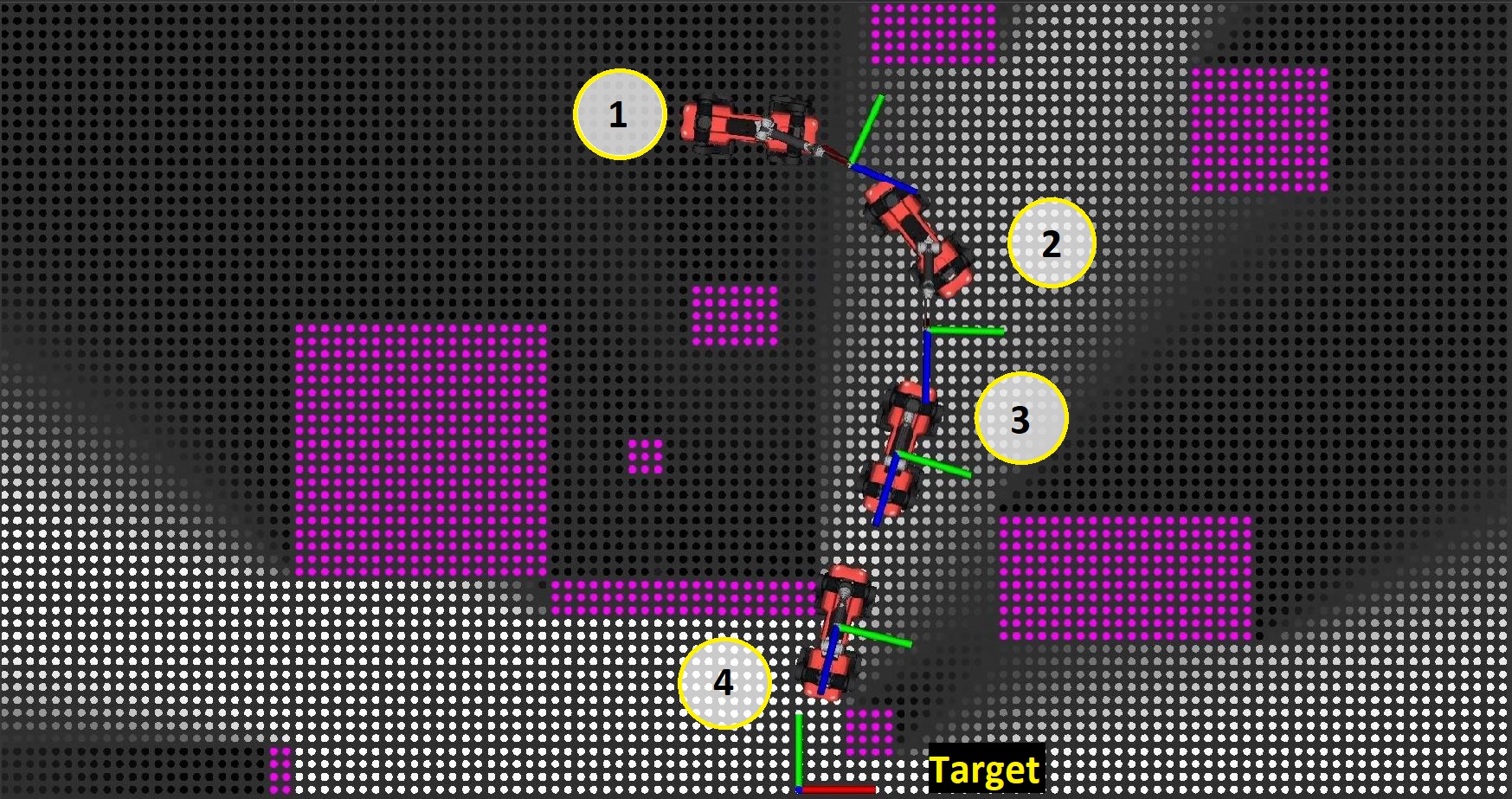}
    \end{subfigure}\hfil
    \caption{Multiplicity illustrations of our physics-based simulations in the two scenes introduced in Fig.~\ref{fig:sim_scenes}. The solver solution in each simulation is reflected through 4 different snapshots of the robot. The target is tagged. Violet points correspond to obstacles at the level of the \emph{ee}.}\label{fig:sim_scenes_animation}
\end{figure}

The final contribution of this paper is two experiments running in a physics-based simulation environment, Gazebo, that validate our motion planning and control pipeline. The MPC horizon is set to 1.0s, and the relevant penalties arise from the actuator inputs, the visibility and orientation costs, height tracking, and the actuator speed, position, and torque limits. Other regularization costs and gait-related constraints are active during this process. Since MPC and mapping run on two different computers, our visibility cost only introduces an additional lookup computational cost, meaning no additional overhead cost to motion planning pipeline when compared to classical methods. We also provide the simulation with a given occupancy grid built based on simulated obstacles in the environment. The grid resolution we adopt is 1000 voxels per cubed meter. In Fig.~\ref{fig:sim_scenes}, we illustrate the top view of each of the two scenes in which we carry out the simulation. In both illustrations, a slice of the \emph{shadow field} at the \emph{ee} level is shown in gray-scale with the same intensity scale as previously introduced for Fig.~\ref{fig:shadow_field_hw_test}. Umbras and penumbras described there can also be noted here. As shown in Fig.~\ref{fig:sim_scenes}, we make sure that the \emph{ee} starts in an occluded position. The robot must then plan a least-shadowy path for the \emph{ee} position that leads it to establish line of sight with the light position. 

In Fig.~\ref{fig:sim_scenes_animation}, we illustrate the robot's motion and its \emph{ee} frame pose in a top view. In the first simulation, the four snapshots show the robot: in its starting occluded position, sliding along the surface of an occluding obstacle so as to circumnavigate it, extending its arm towards regions with higher visibility, and in its steady state having reached maximum visibility. The \emph{ee} also fully locks onto the target, achieving an error complement of 0.995. Similar behaviour can be seen in the second simulation where we block the solution of the first simulation by enlarging the obstacle facing the target and we move the obstacle on its right closer to it. ALMA peaks its \emph{ee} through the narrow passage so as to maximize likelihood of visibility. As an interesting emergent behavior, the \emph{ee} avoids collisions with obstacles since their \emph{shadow field} values are zero. This behavior may be extended to other robot frames.

\section{Conclusion}

In this work, we proposed an MPC formulation based on visibility constraints. We augmented our motion planning cost function with a penalty maximizing relaxed log-likelihood of visibility probability. We introduced a probabilistic \emph{shadow field} that quantifies visibility probability based on the occupancy map of the scene. We validated the quality of this map in simulation and hardware. We further discussed the computational and storage complexities of our \emph{shadow field} mapping and showcased its computational efficiency for onboard applications by real-time mapping on ALMA hardware. A comparison between hard and soft shadows for one 2D implementation shows the extent and accuracy of our approximation. Last but not least, we demonstrated the validity of our proposed MPC formulation for motion planning and dynamic occlusion avoidance in simulation for ALMA. Our proposed solution inherits limitations from gradient-based methods. Moreover, our target needs to be properly mapped in a way that won't block its own line of sight with the robot. This problem must be tackled in future work as we fully validate the entire pipeline on hardware.

\bibliographystyle{bibtex/IEEEtran} 
\bibliography{bibtex/IEEEabrv, bibtex/bibliography}
 
\end{document}